\pdfoutput=1

\documentclass[11pt]{article}

\usepackage[preprint]{acl}

\usepackage[T1]{fontenc}

\usepackage[utf8]{inputenc}

\usepackage{times}
\usepackage{latexsym}
\usepackage{microtype}

\usepackage{inconsolata}

\usepackage{graphicx}

\usepackage{xcolor}
\usepackage{colortbl}
\usepackage{booktabs}
\usepackage{amssymb}
\usepackage{amsmath}
\usepackage{fixltx2e}
\usepackage{listings}
\usepackage{xcolor}
\usepackage{graphicx}
\usepackage{multirow}
\usepackage{array}
\usepackage{algorithm}
\usepackage{algpseudocode}
\usepackage{soul}

\title{Learning to Insert \texttt{[PAUSE]} Tokens for Better Reasoning}

\author{Eunki Kim$^\ast$~~~~ Sangryul Kim$^\ast$~~~~ James Thorne \smallskip \\
        KAIST AI\\ 
        \texttt{\{eunkikim, sangryul, thorne\}@kaist.ac.kr}\\}

\begin{document}
\maketitle
\begingroup\def\thefootnote{$\ast$}\footnotetext{Equal contribution.}\endgroup

\begin{abstract}
To enhance reasoning capabilities, previous works have explored incorporating special-purpose tokens into the training process. These strategies strengthen the learning mechanism of transformer-based large language models (LLMs). Building on prior research, in which inserting dummy tokens consecutively just before reasoning steps can enhance effectiveness, we introduce a novel approach termed \textbf{D}ynamic \textbf{I}nserting Tokens \textbf{T}raining \textbf{(DIT)}. Our method identifies positions within sequences where model confidence is lowest according to token log-likelihood. Strategically inserting \texttt{[PAUSE]} tokens on these positions bolsters the model's predictive capabilities for subsequent tokens. Experimental results across diverse datasets and models, from the 2.7B model to the 8B model, demonstrate that DIT consistently outperforms traditional fine-tuning and previous token insertion methods. With this simple yet effective method, we achieve accuracy gains of up to 4.7\%p on GSM8K, 3.23\%p on AQUA-RAT, and pass@1 improvements of up to 3.4\%p on MBPP datasets. Our work shows a model-based, dynamic approach rather than a heuristic one, thereby broadening the scope of research in reasoning\footnote{Code and datasets are available at \url{https://github.com/xfactlab/acl2025-dit}}.

\end{abstract}

\begin{figure}[t]
    \centering
    \includegraphics[width=1.0\linewidth]{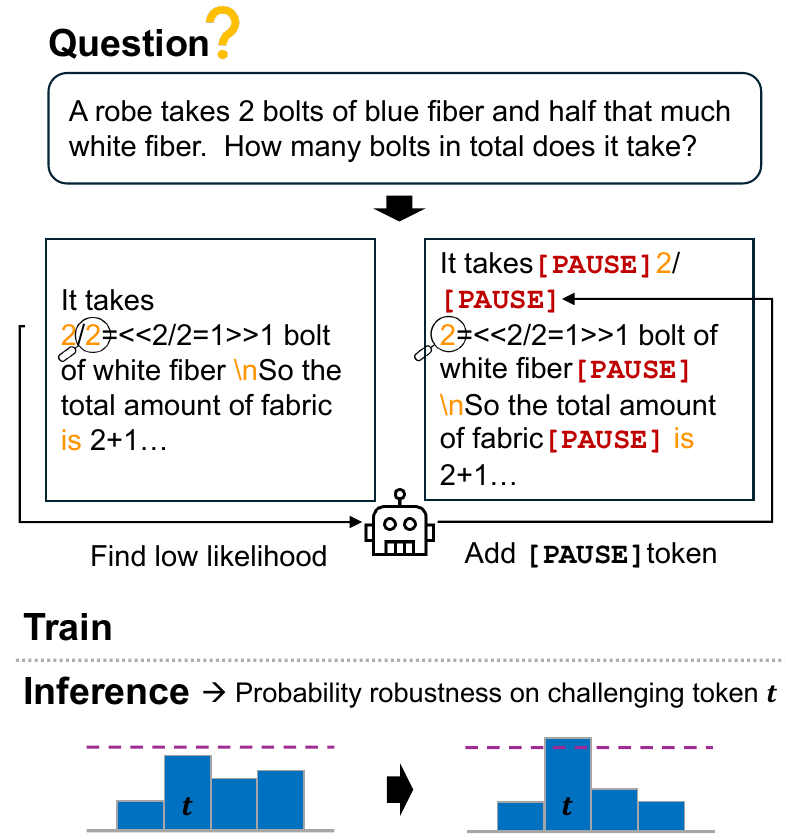}
    \caption{We propose DIT, a log-likelihood-based \texttt{[PAUSE]} token inserting methodology. We suggest that learning with \texttt{[PAUSE]} token insertion using this method will result in a probability boost for low-probability tokens during inference.}
    \label{fig:abstract_img}
\end{figure}

\section{Introduction}

The primary mechanism in the text generation of recent large language models (LLMs) with the decoder-based transformer architecture \cite{vaswani2017attention, Radford2018ImprovingLU} involves autoregressive calculation. These models evaluate the probability of the best subsequent token, determined by the sequence of preceding tokens \cite{vaswani2017attention,radford2018improving,touvron2023llama}. This next token prediction (NTP) paradigm has enabled remarkable advances in language understanding and generation, with performance scaling predictably as model size and training data increase, by scaling laws \cite{brown2020language, kaplan2020scaling}. However, these improvements come at the cost of substantial computational and energy demands, posing significant challenges for efficiency and sustainability \cite{samsi2023words, faiz2023llmcarbon}. 

To effectively adapt LLMs for specific tasks, additional reasoning steps generally support adaptation, using approaches such as the Chain of Thought \citep[CoT]{wei2022chain}, which can be trained as a subsequent objective \cite{ho2022large, shridhar2022distilling}.
While effective, training decoder-only LLMs with NTP loss based on previous input tokens does not always yield optimal results \cite{bachmann2024pitfalls, dziri2024faith}. For instance, LLMs often fail on surprisingly trivial problems in mathematical reasoning tasks \cite{dziri2024faith, ahn2024large}. To address such limitations, recent studies have explored mechanisms for adding dynamic computational budgets for transformer architectures. For instance, the Universal Transformer \cite{dehghani2018universal} adapts model depth based on input complexity, while speculative decoding \cite{leviathan2023fast} offloads demanding reasoning to larger expert models.

Recently, \citet{goyal2024think} have shown that inserting dummy tokens during pre-training and decoding can act as a mechanism for introducing additional computational steps. Building on prior studies \cite{sukhbaatar2019augmenting, xue2023adaptive, darcet2023vision}, these dummy tokens strategically delay the output of decoder-only models, creating additional computational pathways in the attention layers and improving performance. However, this method relies on the stochastic inclusion of the dummy tokens during pre-training, followed by their fixed placement (depending on hand-crafted rules) during fine-tuning. 

Rather than relying on heuristic methods, we propose a model-based approach for dummy token insertion, focusing solely on fine-tuning. Our novel method identifies points with low token log-likelihoods, interpreted as regions of model uncertainty, and subsequently inserts dummy tokens (denoted as \texttt{[PAUSE]}) immediately preceding these positions. This approach is based on previous works that set token log probability as an indicator of the model's confidence score \cite{nguyen-oconnor-2015-posterior, jiang-etal-2023-active, kadavath2022language}.
We conduct experiments on two reasoning tasks: mathematical reasoning and code reasoning, using three datasets (GSM8K \cite{cobbe2021training}, AQUA-RAT \cite{ling-etal-2017-program}, and MBPP \cite{austin2021program}). Our results demonstrate the broad applicability and effectiveness of the proposed method, yielding consistent performance gains. Specifically, our method shows performance improvements over existing methods, with gains of up to 4.7\%p on GSM8K, 3.23\%p on AQUA-RAT in accuracy, and a 3.4\%p increase in pass@1 on MBPP.

We demonstrate that by incorporating \texttt{[PAUSE]} tokens, the model can focus more on challenging tokens with higher loss, resulting in a higher training signal for harder tokens, improving training and inference. By using dummy tokens as a contributor to the loss term for succeeding tokens, we highlight the importance of learning difficult tokens, supported by an in-depth analysis of token probability. Furthermore, our experimental analysis reveals how \texttt{[PAUSE]} tokens enhance the probability robustness of outliers, solidifying their role in improving overall model performance.

To summarize, our contributions are as follows:
\begin{itemize}
\item{We propose a novel method for dynamically inserting dummy tokens into sequences based on token likelihood (DIT).}
\item{We demonstrate performance improvements with our approach, achieving up to 4.25\%p higher accuracy on GSM8K, 3.23\%p on AQUA-RAT, and up to 3.4\%p improvement in pass@1 on MBPP.}
\item{By using dummy tokens as a loss contributor, we highlight the importance of learning challenging tokens, supported by an in-depth analysis of token probability.}
\end{itemize}

\section{\texttt{[PAUSE]} Token}

\citet{goyal2024think} introduce a method for pre-training models with dummy tokens (denoted as the \texttt{<pause>} tokens) that enables arbitrary allocation of additional computation at run-time through the rule-based insertion of the \texttt{<pause>} tokens at inference.
This approach involves using $(K+M)$ vectors instead of just the $K$ existing vectors when generating the $(K+1)^{th}$ token by adding $M$ \texttt{<pause>} tokens to each dataset instance. In their study, the authors introduce a method for incorporating \texttt{<pause>} tokens into a pre-training sequence \( y_{1:N} \). They insert multiple \texttt{<pause>} tokens at random positions represented as \( M_{pt} \), throughout the sequence to get a \texttt{<pause>}-injected sequence, \( \Tilde{y}_{1:N+M_{PT}} \). The model is trained using the next token prediction (NTP) loss while disregarding the loss associated with predicting the \texttt{<pause>} tokens. The positions where the next token is a \texttt{<pause>} token are denoted by \( S_{ignore} = \{k : \Tilde{y}_{k+1} = \texttt{<pause>} \} \). The pre-training loss for a decoder-only language model \( f_{\theta} \) is formulated as follows:
\begin{equation}\label{eqn:pause_pt}
    \begin{gathered}
    \mathcal{L}_{PausePT}(f_\theta, \Tilde{y}_{1:N+M_{PT}}) \\ = \sum^{N+M_{PT}-1}_{k=1, k \notin S_{ignore}} \mathcal{L}_{CE}(\Tilde{y}_{k+1}, f_{\theta}(\Tilde{y}_{1:k})),
    \end{gathered}
\end{equation} where \(\mathcal{L}_{CE}\) denotes the cross-entropy loss. For fine-tuning, given a prefix \( y_{1:N} \) annotated with target \( \mathbf{t}_{1:T} \), they append a series of \texttt{<pause>} tokens (denoted as \( M_{ft} \)) to the end of the prefix, resulting in a \texttt{<pause>}-appended prefix \( \Tilde{y}_{1:N+M_{ft}} \). Using this \texttt{<pause>}-appended prefix, the model is fine-tuned with the NTP loss on the target \( \mathbf{t}_{1:T} \), minimizing the loss as follows:
\begin{align}\label{eqn:pause_ft}
\sum_{k=0}^{T-1} \mathcal{L}_{\rm CE}(\mathbf{t}_{k+1} , f_{\theta}([\Tilde{y}_{1:N+M_{\rm ft}}, \mathbf{t}_{1:k}])  ),
\end{align} where $[ \cdot ]$ denotes the concatenation operation. Authors suggest that inserting \texttt{<pause>} tokens during pre-training and their subsequent utilization during fine-tuning for downstream tasks can create additional computational pathways within the attention layers and the associated MLP layers. This approach has demonstrated significant performance enhancements across various downstream tasks.

However, they show that merely using \texttt{<pause>} tokens for fine-tuning without its prior insertion during pre-training does not result in substantial performance gains. Additionally, in reasoning tasks such as GSM8K, the performance deteriorates when the \texttt{<pause>} token insertion is exclusively implemented during the fine-tuning stage. To address these limitations, we propose a targeted approach for the effective placement of dummy tokens (denoted as \texttt{[PAUSE]}) in reasoning tasks and clarify their role in model training. Previous studies have primarily focused on the pre-training phase, which is resource-intensive. Instead, our research demonstrates that significant performance gains can be achieved by strategically inserting \texttt{[PAUSE]} tokens during the fine-tuning phase. Unlike the arbitrary insertion of \texttt{[PAUSE]} tokens during pre-training, our method identifies optimal positions during fine-tuning, enhancing performance even in models not pre-trained with them.

\section{Methods}

\begin{figure*}[t]
    \centering
    \includegraphics[width=0.9\textwidth]{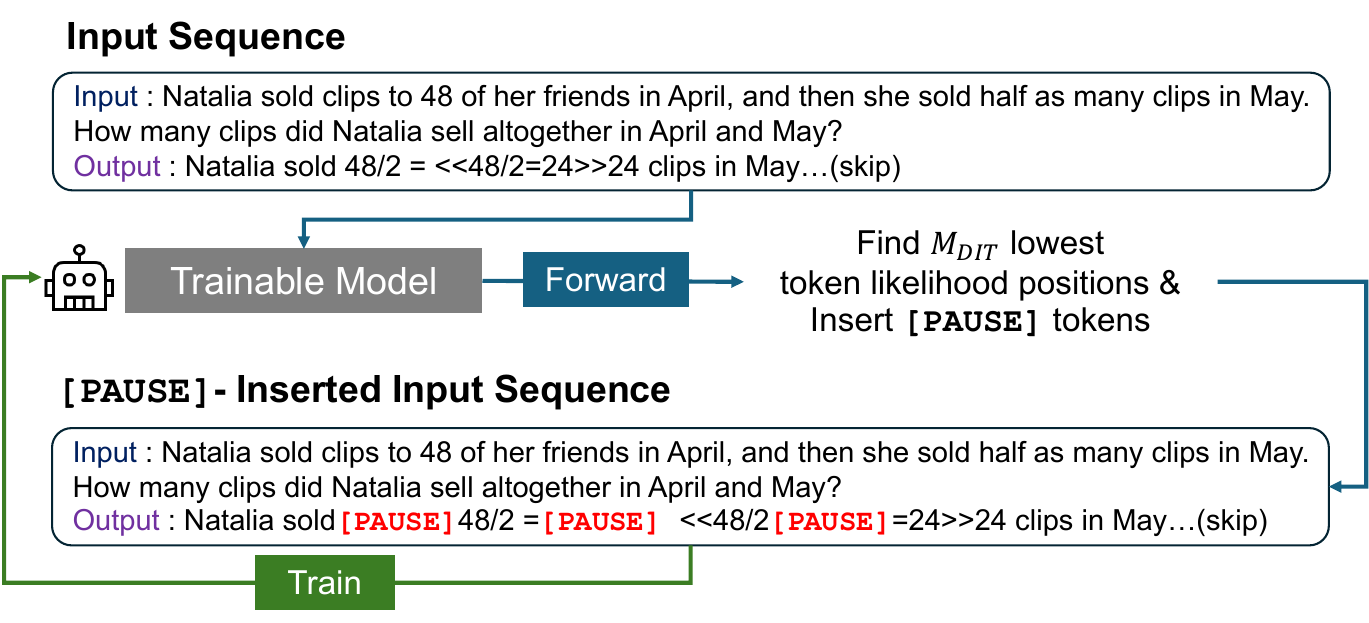}
    \caption{We illustrate the DIT training process, which is divided into two steps. In the first step, we perform a forward pass to identify $M_{DIT}$ positions with low log-likelihood in the model's input. In the second step, we insert \texttt{[PAUSE]} tokens at those positions and train the model using the modified data.}
    \label{fig:train_img}
\end{figure*}

\subsection{Training with \texttt{[PAUSE]} Token}
\label{method:train}

To incorporate the \texttt{[PAUSE]} token during fine-tuning, we first determine the positions of effective insertion. Unlike previous research, our novel method activates \texttt{[PAUSE]} token insertion during the training phase, based on token log-likelihood rather than heuristic or rule-based approaches. Since the inference of the language model starts from the prefix (\textit{e.g.}, prompt, question) and predicts the target (\textit{e.g.}, output), we strategically insert \texttt{[PAUSE]} tokens within the target sequence, which is our primary focus. By executing a forward pass of the model, we compute the log-likelihoods of each token and pinpoint positions where the log-likelihood is lowest in the target sequence.
Subsequently, we train the model using this modified dataset incorporated \texttt{[PAUSE]} tokens. 
At this point, we provide a hyperparameter $M_{DIT}$ to tune the number of \texttt{[PAUSE]} tokens to insert per data instance. The pipeline is described in Figure \ref{fig:train_img}, with an example data.

Consider a sequence \( Y_{1:N} = \{y_1, y_2, \ldots, y_N\} \) that the language model is learning. In the traditional fine-tuning scenario (SFT), the loss for a token \( y_{k+1} \), which ranks within the top $M_{DIT}$ lowest log-likelihoods at time step \( k \), is given by:
\begin{equation}\label{eqn:fines}
\mathcal{L}_{SFT}(f_\theta, y_{k+1}) = \mathcal{L}_{CE}(y_{k+1}, f_\theta(y_{1:k})),
\end{equation}where $\mathcal{L}_{CE}$ denotes the cross-entropy loss. The overall loss for fine-tuning on the sequence \( Y_{1:N} \) is given by:
\begin{equation}\label{eqn:fines_whole}
        \mathcal{L}_{SFT}(f_\theta, Y_{1:N})  = \sum^{N-1}_{k=1} \mathcal{L}_{CE}(y_{k+1}, f_\theta(y_{1:k})).
\end{equation}Conversely, in the DIT (\textbf{D}ynamic \textbf{I}nserting Tokens \textbf{T}raining) scenario, where the \texttt{[PAUSE]} tokens are inserted, the loss becomes:
\begin{equation}\label{eqn:dit}
    \begin{gathered}
        \mathcal{L}_{DIT}(f_\theta, \Tilde{y}_{k+1}) = \mathcal{L}_{CE}(\Tilde{y}_{k+1}, f_\theta([\Tilde{y}_{1:k}, \texttt{[PAUSE]}])),
    \end{gathered}
\end{equation}where $\Tilde{Y}_{1:N+M_{DIT}} = \{\Tilde{y}_1, \Tilde{y}_2,\ldots, \Tilde{y}_{N+M_{DIT}}\}$ denotes pause-injected sequences with $M_{DIT}$ \texttt{[PAUSE]} tokens inserted.
The overall loss for DIT on the sequence \( \Tilde{Y}_{1:N+M_{DIT}} \) is given by:

\begin{equation}\label{eqn:dit_whole}
    \begin{gathered}
        \mathcal{L}_{DIT}(f_\theta, \Tilde{Y}_{1:N+M_{DIT}})\\  = \sum^{N+M_{DIT}-1}_{k=1, k \notin S_{ignore}} \mathcal{L}_{CE}(\Tilde{y}_{k+1}, f_\theta(\Tilde{y}_{1:k})).
    \end{gathered}
\end{equation}
This is similar to equation \ref{eqn:pause_pt}, but in the DIT scenario, \( S_{ignore} \) is determined based on token log-likelihood rather than random selection.

\subsection{Heuristic Findings}
We provide justifications and an overview of preliminary experiments that elucidate the rationale behind the development of DIT. In our prior experiments, we leverage a fine-tuned model to pinpoint optimal positions for inserting \texttt{[PAUSE]} tokens. During the token generation for a given input, we calculate the log probabilities of each token and pinpoint the positions of the lowest tokens. Drawing on prior research that underscores log probability as a crucial indicator of a model's confidence and its knowledge gaps \cite{jiang-etal-2023-active, kadavath2022language}, we identify strategic positions for inserting \texttt{[PAUSE]} tokens to enhance downstream task performance.

Interestingly, our observations reveal that log probabilities at specific points within each task tend to be low across experimental datasets. Training our models by inserting \texttt{[PAUSE]} tokens before these low-probability tokens in the training datasets results in noticeable performance improvements. This finding suggests that the model's reasoning or generation confidence diminishes according to certain reasoning patterns, and the \texttt{[PAUSE]} token aids in learning these challenging tokens. By inserting \texttt{[PAUSE]} tokens at these critical locations, we aim to validate their effectiveness through changes in downstream metrics. 
However, this finding is based on heuristic and task-specific methods for determining the positions to insert \texttt{[PAUSE]} tokens. From this point, we report on automated methods to identify optimal positions for \texttt{[PAUSE]} token insertion and train the model accordingly.
We report the results of these experiments in Appendix \ref{apdx:apriori}.

\section{Experiments}
\subsection{Comparison against other works}

Our method is dynamically inserting \texttt{[PAUSE]} tokens based on the calculated token likelihood, rather than relying on statically predetermined positions. We validate our method through comparison against the following  methods and baselines:
\begin{itemize}
    \item {\textbf{D}ynamic \textbf{I}nserting Tokens \textbf{T}raining (\textbf{DIT})}: Our method (\S\ref{method:train}) dynamically inserts a \texttt{[PAUSE]} token in front of any token that is among the bottom $M_{DIT}$ tokens in terms of the likelihood of the generated output token.
    \item {Supervised Fine-Tune (\textbf{SFT}): A model that is fine-tuned using only the given downstream data without inserting any tokens. This serves as the baseline experiment for all comparisons in our study.}
    \item {Random (\textbf{RAN})}: Insert \texttt{[PAUSE]} tokens at random positions between output tokens for a given input sequence in the training set. 
    \item {Append (\textbf{APPD})}: Appending \texttt{[PAUSE]} tokens at the end of the input in the given training set has been recognized as the best performing one in previous studies \cite{goyal2024think}. During inference, we also append \texttt{[PAUSE]} tokens at the end of the prefix, following established practices.
    \item {After All Words (\textbf{AAW}}): \citet{herel2024thinking} suggests inserting so-called ``thinking tokens(\texttt{<T>})'' after each word in a sentence to introduce extra computational steps within the hidden layers, resulting in improved perplexity. To compare our method, we utilize this idea with the same setting, but we replace the thinking tokens with \texttt{[PAUSE]} tokens.
\end{itemize}

\begin{table*}[ht]
    \centering
    \normalsize
    \begin{tabular}{cl|ccc}
    \toprule
    \multicolumn{2}{c}{} & \multicolumn{3}{c}{\textbf{Dataset}} \\
    \textbf{Base Model} & \textbf{Method} & \small{\textbf{GSM8K}} & \small{\textbf{AQUA-RAT}} & \small{\textbf{MBPP}} \\ 
        \midrule
        \multirow{5}{*}{Phi-2} &  SFT & 51.63 & 40.15 & \underline{14.0} \\
                        & RAN & \underline{54.20} & 32.03 & 0.0 \\
                        & APPD & 52.38 & \underline{41.43} & 10.4 \\
                        & AAW & 39.04 & 17.03 & 5.4 \\
                        & \textbf{DIT(ours)} & \textbf{56.33} & \textbf{43.38} & \textbf{17.4} \\
        \midrule
        \multirow{5}{*}{Phi-3 Mini} &  SFT & \underline{78.84} & 62.46 & \textbf{32.2} \\
                            & RAN & 75.43 & 48.00 & 17.0 \\
                            & APPD & \textbf{79.45} & \textbf{63.22} & \underline{29.6} \\
                            & AAW & 73.84 & 19.83 & 22.4 \\
                            & \textbf{DIT(ours)} & 78.62 & \underline{62.82} & \textbf{32.2} \\
        \midrule
        \multirow{5}{*}{Llama 3 8B} &  SFT & \underline{65.04} & \underline{70.71} & \underline{12.6} \\
                            & RAN & 60.95 & 54.92 & 9.4 \\
                            & APPD & 63.83 & \textbf{72.18} & 11.0 \\
                            & AAW & 64.06 & 19.25 & 9.2 \\
                            & \textbf{DIT(ours)} & \textbf{66.86} & 69.96 & \textbf{14.0} \\
    \bottomrule
    \end{tabular}
    \caption{\textbf{Impact of Dynamic Inserting Tokens Training (DIT) on Various Datasets.} In this table, we showcase the performance comparison across three base models—Phi-2, Phi-3 Mini, and Llama 3 8B—evaluated on three distinct datasets: GSM8K (accuracy), AQUA-RAT (accuracy), and MBPP (pass@1). The methods evaluated include Supervised Fine-Tune (SFT), Random (RAN), Append (APPD), After All Words (AAW), and our proposed DIT (ours). For each method, the best score is indicated in \textbf{bold}, and the second-best score is \underline{underlined}.}
    \label{tab:dit_result}
\end{table*}

\subsection{Downstream Datasets}

We conduct experiments utilizing datasets designed to assess the reasoning and generative capabilities of the models. These include two renowned mathematical reasoning datasets and one code reasoning dataset. The specific processing and settings are detailed in Appendix \ref{apdx:datasets}.

\paragraph{Math Reasoning}
\textbf{GSM8K} \cite{cobbe2021training} consists of 8.5K diverse grade school math word problems written in linguistic expression containing calculation annotations with special tokens(\textit{e.g.}, `<<',`>>') where arithmetic calculation occurs. \textbf{AQUA-RAT} \cite{ling-etal-2017-program} is a dataset that includes a correct answer and a rationale, a natural language description of the solution for each given math problem, among five possible options. For both datasets, the metric is accuracy, which is determined by whether the final answer, generated through the rationales, is correct or incorrect.

\paragraph{Code Reasoning}
\textbf{MBPP} \cite{austin2021program} is a dataset consisting of an instruction prompt, the corresponding Python function and module to be generated accordingly, and three test cases along with code to verify its functionality. The metric is pass@1, which requires the model to generate one piece of code from the instruction that passes all provided test cases.

\subsection{Models}

In this experiment, we leverage well-known open-source large language models, specifically Phi-2 \cite{microsoft_research_2023}, which comprises 2.7B parameters. Additionally, we include the latest models, Phi-3 Mini (3.8B parameters) \cite{abdin2024phi} and Llama 3 (8B parameters) \cite{llama3modelcard}, as baseline models. This diverse setup allows us to systematically compare the impact of the \texttt{[PAUSE]} token across various models with differing levels of pre-training and parameter counts. Details of experiment settings are described in Appendix \ref{apdx:settings}.

\section{Results and Analysis}

\subsection{Impact of \texttt{[PAUSE]} Token}
\label{sec:impact}

The experimental results, summarized in Table \ref{tab:dit_result}, clearly demonstrate the efficacy of the \texttt{[PAUSE]} token insertion, especially using the DIT method. Notably, DIT consistently shows optimal performance across multiple datasets and model configurations.

For the Phi-2 model, DIT outperforms all other methods, achieving performance improvements of 4.7\%p on GSM8K, 3.23\%p on AQUA-RAT, and 3.40\%p on MBPP compared to SFT. These result indicate that our approach of inserting \texttt{[PAUSE]} tokens at strategically identified positions significantly enhances performance. Similarly, with the Phi-3 Mini model, DIT achieves a 0.36\%p performance improvement on AQUA-RAT compared to SFT. In addition, DIT outperforms other token-insertion methodologies on MBPP. This further validates the effectiveness of our method across different model sizes and pre-training levels. For the Llama 3 8B model, DIT achieves performance improvements of 1.82\%p on GSM8K and 1.4\%p on MBPP compared to the baseline. 
For Llama 3 on AQUA-RAT and MBPP, the performance of DIT slightly decreases compared to SFT. This may be due to the application of instruction format, not generally used for Llama 3, and the generalization of hyperparameters (\textit{e.g.}, learning rates and epochs) for fine-tuning, which may not be optimal. 

\begin{figure}[t]
    \raggedright\includegraphics[width=1.0\linewidth]{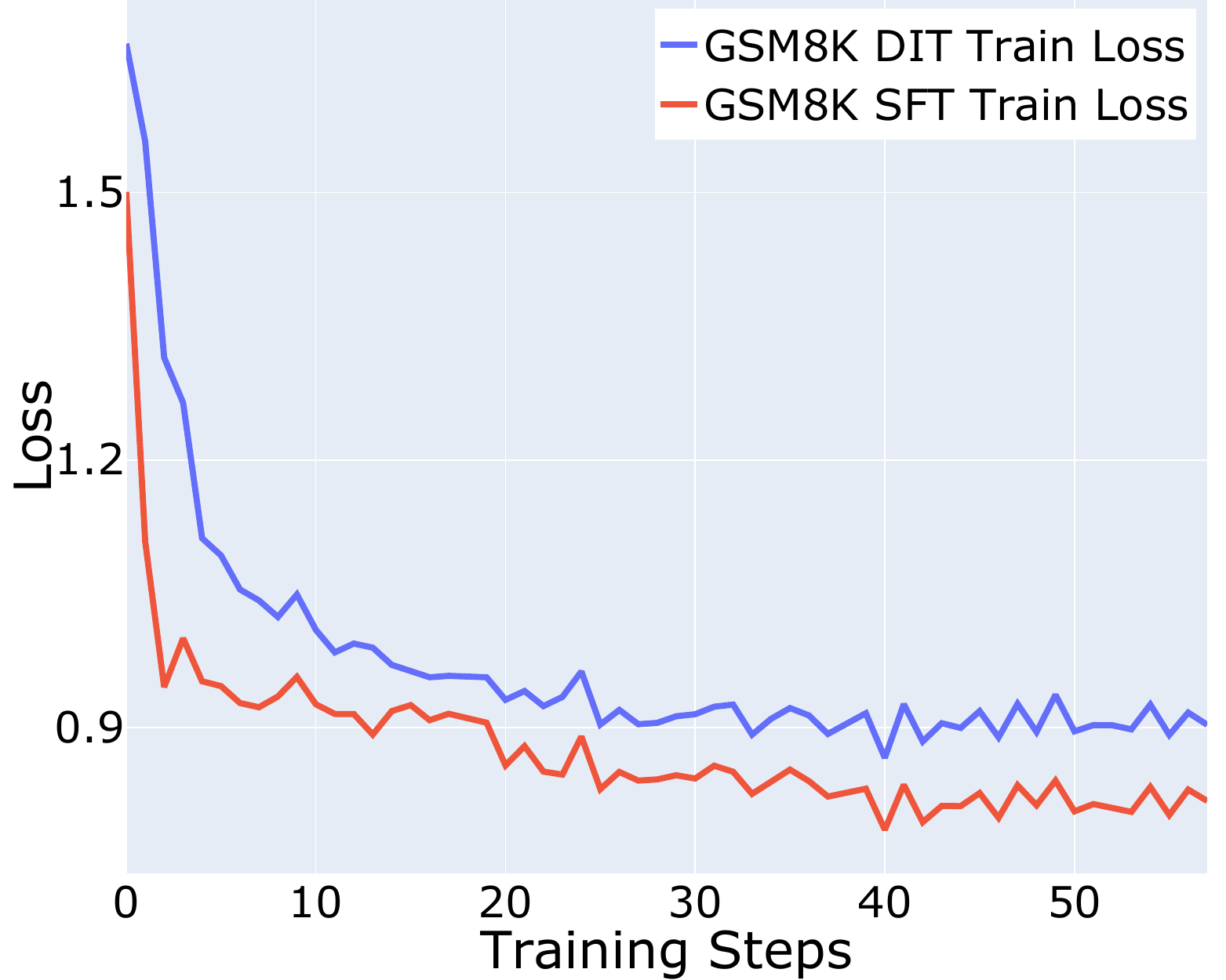}
    \caption{\textbf{Train Loss Comparison between SFT and DIT (ours) on GSM8K Dataset.} The DIT approach results in higher training loss compared to SFT, even when excluding losses associated with \texttt{[PAUSE]} tokens.}
    \label{fig:loss}
    \vspace{-1.0em}
\end{figure}

To summarize, across the nine experimental settings comparing our method with the baseline SFT and other \texttt{[PAUSE]} token-inserting methods, DIT demonstrated superior performance in six cases, achieving the highest scores. Additionally, in one case, DIT secured the second-best performance, further emphasizing the effectiveness and robustness of our approach. Although it does not surpass all other methods on every dataset, it remains highly competitive, underscoring the robustness of the \texttt{[PAUSE]} token strategy. In conclusion, DIT consistently demonstrates superior performance across various datasets, affirming that strategically inserting \texttt{[PAUSE]} tokens can significantly enhance model efficacy in reasoning tasks.

\begin{figure}[h]
    \centering
    \includegraphics[width=1.0\linewidth]{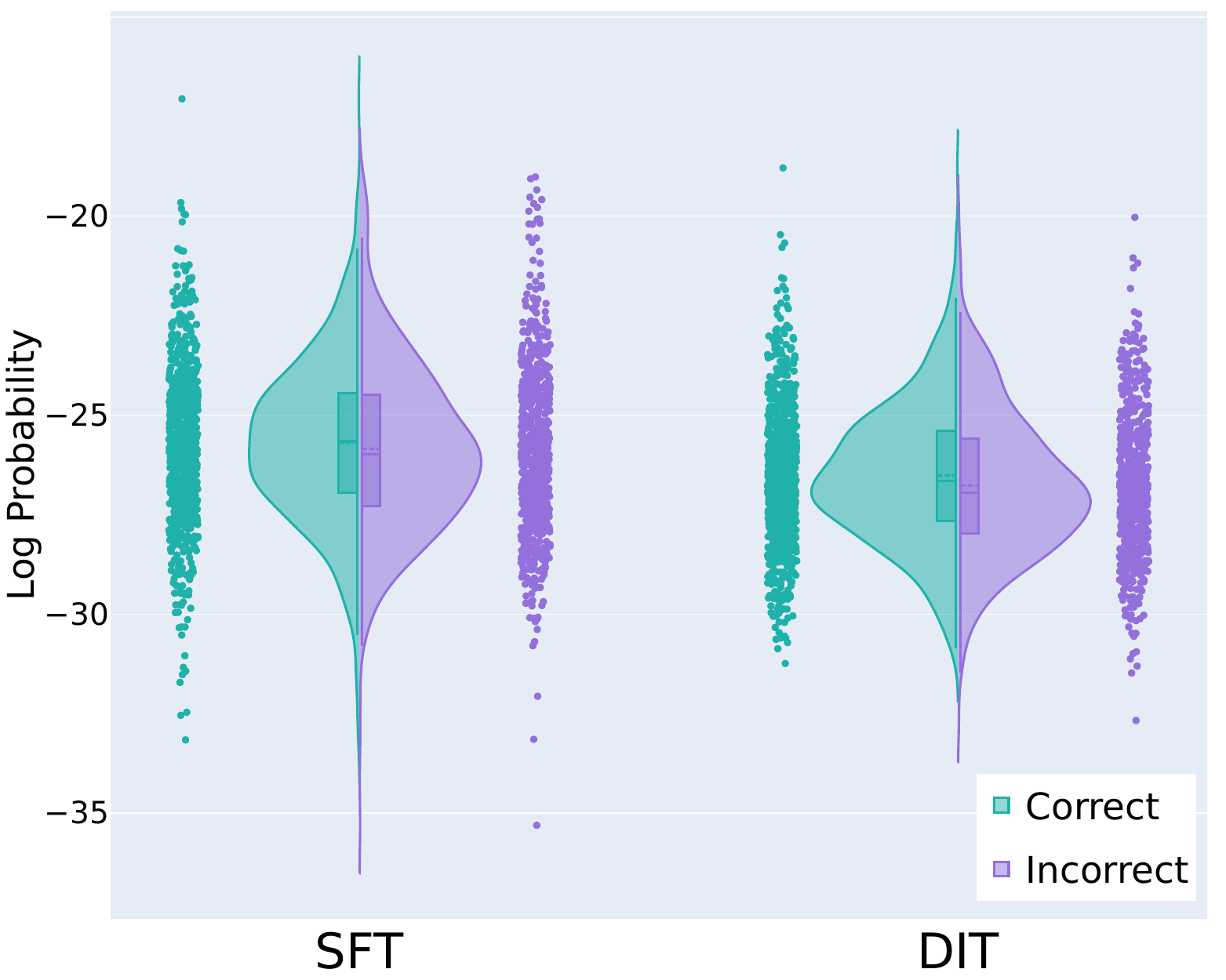}
    \caption{\textbf{Comparison of the lowest 20 log probabilities for tokens between SFT and DIT (ours) on the GSM8K dataset.} The violin plots illustrate that Dynamic Insertion of \texttt{[PAUSE]} Tokens (DIT) results in higher log probabilities for tokens initially ranked with low probabilities, indicating enhanced model confidence in predicting challenging tokens.}
    \label{fig:logp}
    \vspace{-1.0em}
\end{figure}

\subsection{Role of \texttt{[PAUSE]} Token in Training}
\label{sec:role}
We investigate the impact of inserting \texttt{[PAUSE]} tokens into sequences during language model training. As shown in Figure \ref{fig:loss}, our results indicate that irrespective of the model and dataset, the insertion of \texttt{[PAUSE]} tokens leads to an increase in losses. The observed increase in losses, where \( \mathcal{L}_{DIT}(f_\theta,\Tilde{Y}_{1:N+M_{DIT}}) \) exceeds \( \mathcal{L}_{SFT}(f_\theta, Y_{1:N}) \) suggests that the presence of the \texttt{[PAUSE]} token complicates the model's ability to predict \( \Tilde{y}_{t+1} \) based on the preceding tokens up to step \( t \).

By strategically inserting a \texttt{[PAUSE]} token at a position where the log-likelihood of \( \Tilde{y}_{t+1} \) is already among the lowest in the sequence, the log-likelihood further decreases, increasing in the loss term. This method effectively heightens the loss for tokens that are challenging to predict based on previous tokens, thereby encouraging the model trained more robustly. Previous studies have highlighted that dynamically adjusting the loss term for challenging instances can mitigate performance degradation \cite{jiang2020tldr, lin2017focal, fernando2021dynamically, wang2022language}. Consequently, DIT induces a more significant shift in the model's parameters compared to standard fine-tuning, enhancing the model's fit to the dataset. This improved alignment likely accounts for the observed performance improvements.

\subsection{\texttt{[PAUSE]} token with token probability}

In this section, we delve into the impact of \texttt{[PAUSE]} token insertion on the performance of downstream tasks, focusing on token log probability. By strategically placing \texttt{[PAUSE]} tokens where model confidence is low during training, we aim to bolster the model's certainty in selecting subsequent tokens, thereby enhancing performance in downstream tasks. This expectation is substantiated by significant performance improvements across various domains, as demonstrated in Table \ref{tab:dit_result} and \S\ref{sec:impact}. To understand the changes during inference, we contrast SFT and DIT methodologies using the GSM8K dataset, depicted in Figure \ref{fig:logp}. Here, the y-axis represents the log probability, with the x-axis contrasting SFT and DIT. Each violin graph showcases correct predictions in cyan on the left and incorrect predictions in violet on the right, with actual log probability values displayed on both sides. These values represent the averages of the lowest 20 log probabilities for tokens in each test dataset, facilitating observation of alterations in low log probability tokens.

As shown in Figure \ref{fig:logp}, DIT exhibits a shortened long tail compared to the SFT method, indicative of heightened log probability. Moreover, the data appears more clustered around the mean, signaling reduced variance. Although overall log probabilities for each data point may slightly decrease due to \texttt{[PAUSE]} token insertion, the probabilities of previously low-confidence tokens experience an upsurge. Essentially, DIT training, facilitated by \texttt{[PAUSE]} token insertion before low log-likelihood tokens, enables the model to prioritize learning challenging tokens. Consequently, while the overall average probability may decrease due to task-irrelevant \texttt{[PAUSE]} tokens, we can see an increase in the average probability of outliers, ultimately leading to performance enhancement.

\subsection{Analysis for the Loss Calibration}

Here, we examine the impact of \texttt{[PAUSE]} tokens on loss calibration, a key factor in ensuring stable learning, especially in models dealing with complex or imbalanced data \cite{mukhoti2020calibrating, tao2023dual}. In our approach, the \texttt{[PAUSE]} token insertion plays a critical role in this process. These tokens act as a temporal regularization, inducing the model to slow down and pay closer attention to challenging tokens that otherwise might be overlooked.

This is similar to the principles behind the focal loss \citep{lin2017focal}, where the loss function is adjusted to emphasize challenging examples of those with low likelihood while down-weighting easier ones. Focal loss adds relatively more weight for misclassified instances by dynamically scaling cross-entropy (CE) loss depending on model confidence. Whereas the conventional CE objective is akin to \(-\log p_t\), FL scales the objective as \( -(1-p_t)^\gamma\log p_t\). This focal loss objective has been adapted for reducing bias in tasks such as natural language inference (NLI) by dynamically reducing instance weighting based on undesirable properties during training \citep{karimi-mahabadi-etal-2020-end}.

Within the context of DIT, the \texttt{[PAUSE]} token is previously unseen and will be associated with a high loss. Given that \(\mathcal{L}_{CE}(\Tilde{y}_{k+1}, f_\theta([\Tilde{y}_{1:k}, \texttt{[PAUSE]}])) > \mathcal{L}_{CE}(\Tilde{y}_{k+1}, f_\theta(\Tilde{y}_{1:k}))\), inserting \texttt{[PAUSE]} tokens with DIT dynamically increases the loss for low-confidence tokens. This elevated loss signal prompts the model to focus more attention on these uncertain tokens, enhancing its ability to learn from challenging contexts. As a result, the models become more robust and show improved performance in downstream tasks, as demonstrated empirically in Section~\ref{sec:role}.

The effect of \texttt{[PAUSE]} tokens can be likened to the NoiseBoost \cite{wu2024noiseboost} method, where perturbations are introduced to balance attention across different modalities. Just as NoiseBoost prevents over-reliance on specific inputs by distributing attention more evenly, \texttt{[PAUSE]} tokens help recalibrate the model's focus during training. This recalibration stabilizes the learning process by preventing the model from becoming overly confident in easy predictions or neglecting challenging tokens, thereby enhancing generalization.

\section{Ablations}

In this section, we conduct an ablation study on the number of \texttt{[PAUSE]} tokens and the consecutive insertion lengths, which we consider important in our experimental setup. Using the GSM8K dataset, which shows the most significant performance improvement over the baseline setting SFT, we perform the ablation study on the DIT model based on the Phi-2 model.

\paragraph{The number of \texttt{[PAUSE]} token}
As detailed in Appendix \ref{apdx:settings}, we insert a total of five \texttt{[PAUSE]} tokens during training. A critical consideration is how many low log-likelihood tokens should have \texttt{[PAUSE]} tokens inserted before them to effectively modify their likelihood values. In addition to our initial setting of five tokens, we experimented with inserting 1, 3, 7, and 10 \texttt{[PAUSE]} tokens under the same conditions.

\begin{figure}[t]
    \centering
    \includegraphics[width=1.0\linewidth]{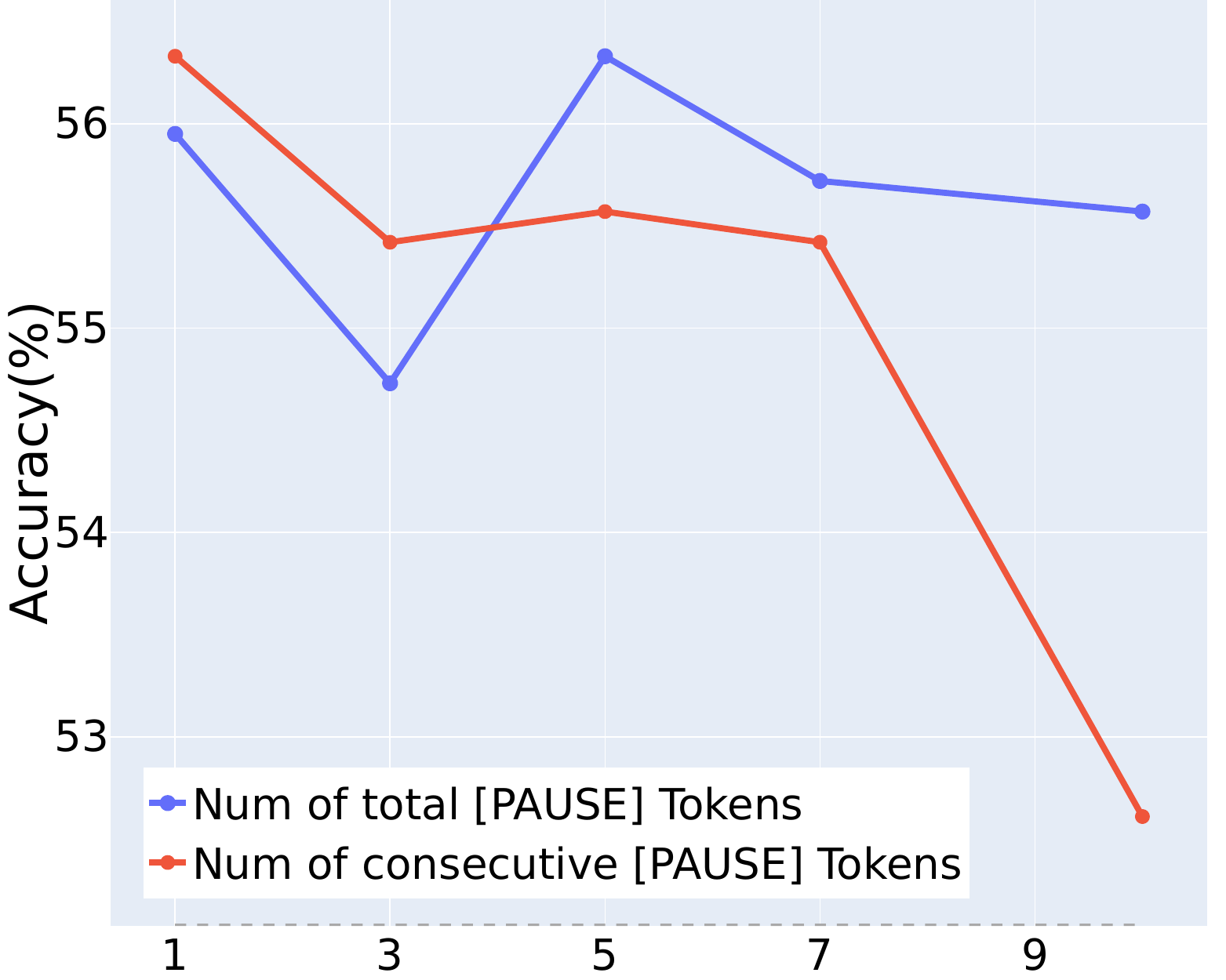}
    \caption{\textbf{Impact of varying the number of \texttt{[PAUSE]} tokens on model performance.} The graph shows that inserting five \texttt{[PAUSE]} tokens achieves optimal performance improvement for the Phi-2 model on the GSM8K dataset.}
    \label{fig:ablation}
\end{figure}

The results, depicted in the blue graph in Figure \ref{fig:ablation}, indicate that our choice of five \texttt{[PAUSE]} tokens is optimal, yielding the highest performance improvement for the Phi-2 model base using the GSM8K dataset. The graph also shows no consistent pattern suggesting that a higher or lower number of \texttt{[PAUSE]} tokens is inherently better. Notably, across all configurations, Dynamic Insertion of \texttt{[PAUSE]} Tokens (DIT) consistently achieves superior scores compared to other methods.

The role of the \texttt{[PAUSE]} token is pivotal: strategically inserting \texttt{[PAUSE]} tokens in front of tokens with a likelihood below a certain threshold can aid in learning challenging tokens. However, as evidenced by the performance degradation of AAW and RAN in Table \ref{tab:dit_result}, inserting \texttt{[PAUSE]} tokens when token log-likelihood is above a certain threshold can potentially disrupt maximizing likelihood or introduce noise. Hence, identifying the optimal number of \texttt{[PAUSE]} tokens to insert is crucial.

\paragraph{Length of \texttt{[PAUSE]} token blocks}
In \citet{goyal2024think}'s work, a series of \texttt{[PAUSE]} tokens is created by inserting $n$ consecutive \texttt{[PAUSE]} tokens, forming token blocks. In contrast, our approach involves inserting a single \texttt{[PAUSE]} token at each designated position. To assess the impact on performance, we experiment with inserting multiple consecutive \texttt{[PAUSE]} tokens at the optimal five positions identified previously. Specifically, we insert 3, 5, 7, and 10 consecutive \texttt{[PAUSE]} tokens at each position. For instance, inserting three consecutive \texttt{[PAUSE]} tokens at each of the five positions results in a total of fifteen \texttt{[PAUSE]} tokens. As shown in Figure \ref{fig:ablation}, the red graph indicates that using a single \texttt{[PAUSE]} token consecutively, as in our approach, is most effective. The performance decreases as the number of consecutive \texttt{[PAUSE]} tokens increases,  proportionally increasing the total number of tokens.

This performance degradation is likely due to the increased delay in the calculation of the next token caused by inserting consecutive \texttt{[PAUSE]} tokens. This finding contrasts with \citet{goyal2024think}'s results, underscoring the significance of the consecutive number of \texttt{[PAUSE]} tokens inserted, which critically impacts performance. Our results suggest that strategically placed single \texttt{[PAUSE]} tokens can significantly improve performance, highlighting the importance of precise placement over the quantity of inserted tokens.

\section{Related Works}
\paragraph{Token Insertion}
Recent decoder-based language models \cite{radford2018improving, microsoft_research_2023, abdin2024phi, llama3modelcard} generate the upcoming token based on the probabilities of previous tokens. This characteristic has been extensively explored in various studies, which have proposed incorporating special tokens to modify the token log-likelihood values. For instance, some have introduced controllable codes in language models \cite{keskarCTRL2019}, while others have explored the removal of tokens during processing in a vision transformer \cite{rao2021dynamicvit}. In addition, the concept of memory transformers has emerged, where dummy tokens are prefixed to token vectors to enable additional trainable memory for storing representations \cite{burtsev2020memory, bulatov2022recurrent}. Despite these advancements, there remains an opportunity for further research on the strategic insertion and removal of tokens in transformer models.

\paragraph{Model Confidence}
Training of LLMs using the transformer's decoder ensures that during inference, generated tokens are based on the highest log probability tokens or methods like beam search to find optimal probabilities within a range \cite{germann-2003-greedy, freitag-al-onaizan-2017-beam}. Hence, when the model generates tokens, a low log probability value for a specific token indicates that the probability differences among token candidates are relatively small. This means that the model is more likely to generate tokens with higher uncertainty. Prior works indicate that this low probability considered as low confidence of the model can indicates a lack of knowledge in model \cite{dong-etal-2018-confidence, jiang-etal-2023-active, kadavath2022language, varshney2022can}. 
Our study, which aims to bolster model confidence on a token-by-token basis, builds upon these foundational insights.

\section{Conclusions}

Previous research \cite{goyal2024think} has underscored the effectiveness of the dummy token in enhancing performance by creating additional computational pathways in attention layers and delaying output generation. Building on this, our study strategically employs \texttt{[PAUSE]} tokens to improve reasoning tasks. We introduced a novel approach where \texttt{[PAUSE]} tokens are dynamically inserted based on token log-likelihood, rather than following fixed rules or relying on randomness. This method targets low-confidence areas, enhancing model stability during training and inference. Our findings demonstrate that inserting \texttt{[PAUSE]} tokens at low log-likelihood positions significantly boosts the model's ability to handle challenging tokens with higher losses and improve the probability robustness of outlier tokens during inference.

\section*{Limitations}
\paragraph{Negative Impact of \texttt{[PAUSE]} token}
During our experiments, we have observed not only the positive outcomes of inserting \texttt{[PAUSE]} tokens but also some negative effects. While DIT generally demonstrates improvements, there are instances where certain datasets shows worse performance compared to straightforward fine-tuning (SFT). Notably, in the Phi-2 GSM8k dataset, where DIT exhibits significant enhancement over SFT, the improvement stems largely from correcting errors introduced by SFT rather than the reverse. Nevertheless, there are cases in the test set where DIT introduce errors absent in the SFT model. These observations prompt further investigation into unintended consequences, including the potential for \texttt{[PAUSE]} tokens to introduce token noise. Addressing these nuances will require additional studies to refine the deployment and optimization of \texttt{[PAUSE]} tokens in future model architectures.

\paragraph{Interpretations Still Obscured Within a Black Box}
In the original work by \citet{goyal2024think}, they mention that it is unclear what changes their method would bring about in practice and offer their own interpretation that a ``wider'' computational pathway induced by the delay could have a positive effect, leading to performance improvements. Our research focuses on optimizing the insertion points of \texttt{[PAUSE]} tokens and automating this process rather than relying on heuristic methods. Therefore, we partially rely on the original interpretation to explain why this approach is effective. From the model's perspective, it is imperative to achieve a more profound comprehension of the effects of the \texttt{[PAUSE]} token.

\section*{Ethics Statement}
In our experiments throughout the paper, we used three models with publicly available weights. Using our method, there is a possibility of obtaining unintended experimental outputs. Additionally, we conduct our experiments using research data available in public repositories. Detailed information about this can be found in the Appendix. It is important to note that all training models and data, as well as our work using them, may produce incorrect outputs if used with improper intent. Also, We acknowledge the use of the LLM, GPT-4o in writing the paper, for grammar and contextual corrections for better expression. Ultimately, all authors rewrite the final version manuscript in their own words.

\section*{Acknowledgments}

This work was partly supported by Institute for Information \& communications Technology Planning \& Evaluation(IITP) grant funded by the Korea government(MSIT) (RS-2019-II190075, Artificial Intelligence Graduate School Program(KAIST)) and Institute for Information \& communications Technology Promotion(IITP) grants funded by the Korea government(MSIT) (RS-2024-00398115, Research on the reliability and coherence of outcomes produced by Generative AI).

\bibliography{Main}

\begin{thebibliography}{47}
\providecommand{\natexlab}[1]{#1}

\bibitem[{Abdin et~al.(2024)Abdin, Jacobs, Awan, Aneja, Awadallah, Awadalla, Bach, Bahree, Bakhtiari, Behl et~al.}]{abdin2024phi}
Marah Abdin, Sam~Ade Jacobs, Ammar~Ahmad Awan, Jyoti Aneja, Ahmed Awadallah, Hany Awadalla, Nguyen Bach, Amit Bahree, Arash Bakhtiari, Harkirat Behl, et~al. 2024.
\newblock Phi-3 technical report: A highly capable language model locally on your phone.
\newblock \emph{arXiv preprint arXiv:2404.14219}.

\bibitem[{Ahn et~al.(2024)Ahn, Verma, Lou, Liu, Zhang, and Yin}]{ahn2024large}
Janice Ahn, Rishu Verma, Renze Lou, Di~Liu, Rui Zhang, and Wenpeng Yin. 2024.
\newblock Large language models for mathematical reasoning: Progresses and challenges.
\newblock \emph{arXiv preprint arXiv:2402.00157}.

\bibitem[{AI@Meta(2024)}]{llama3modelcard}
AI@Meta. 2024.
\newblock \href {https://github.com/meta-llama/llama3/blob/main/MODEL_CARD.md} {Llama 3 model card}.

\bibitem[{Austin et~al.(2021)Austin, Odena, Nye, Bosma, Michalewski, Dohan, Jiang, Cai, Terry, Le et~al.}]{austin2021program}
Jacob Austin, Augustus Odena, Maxwell Nye, Maarten Bosma, Henryk Michalewski, David Dohan, Ellen Jiang, Carrie Cai, Michael Terry, Quoc Le, et~al. 2021.
\newblock Program synthesis with large language models.
\newblock \emph{arXiv preprint arXiv:2108.07732}.

\bibitem[{Bachmann and Nagarajan(2024)}]{bachmann2024pitfalls}
Gregor Bachmann and Vaishnavh Nagarajan. 2024.
\newblock The pitfalls of next-token prediction.
\newblock \emph{arXiv preprint arXiv:2403.06963}.

\bibitem[{Brown et~al.(2020)Brown, Mann, Ryder, Subbiah, Kaplan, Dhariwal, Neelakantan, Shyam, Sastry, Askell et~al.}]{brown2020language}
Tom Brown, Benjamin Mann, Nick Ryder, Melanie Subbiah, Jared~D Kaplan, Prafulla Dhariwal, Arvind Neelakantan, Pranav Shyam, Girish Sastry, Amanda Askell, et~al. 2020.
\newblock Language models are few-shot learners.
\newblock \emph{Advances in neural information processing systems}, 33:1877--1901.

\bibitem[{Bulatov et~al.(2022)Bulatov, Kuratov, and Burtsev}]{bulatov2022recurrent}
Aydar Bulatov, Yury Kuratov, and Mikhail Burtsev. 2022.
\newblock Recurrent memory transformer.
\newblock \emph{Advances in Neural Information Processing Systems}, 35:11079--11091.

\bibitem[{Burtsev et~al.(2020)Burtsev, Kuratov, Peganov, and Sapunov}]{burtsev2020memory}
Mikhail~S Burtsev, Yuri Kuratov, Anton Peganov, and Grigory~V Sapunov. 2020.
\newblock Memory transformer.
\newblock \emph{arXiv preprint arXiv:2006.11527}.

\bibitem[{Cobbe et~al.(2021)Cobbe, Kosaraju, Bavarian, Chen, Jun, Kaiser, Plappert, Tworek, Hilton, Nakano et~al.}]{cobbe2021training}
Karl Cobbe, Vineet Kosaraju, Mohammad Bavarian, Mark Chen, Heewoo Jun, Lukasz Kaiser, Matthias Plappert, Jerry Tworek, Jacob Hilton, Reiichiro Nakano, et~al. 2021.
\newblock Training verifiers to solve math word problems.
\newblock \emph{arXiv preprint arXiv:2110.14168}.

\bibitem[{Dao(2024)}]{dao2024flashattention}
Tri Dao. 2024.
\newblock \href {https://openreview.net/forum?id=mZn2Xyh9Ec} {Flashattention-2: Faster attention with better parallelism and work partitioning}.
\newblock In \emph{The Twelfth International Conference on Learning Representations}.

\bibitem[{Darcet et~al.(2023)Darcet, Oquab, Mairal, and Bojanowski}]{darcet2023vision}
Timoth{\'e}e Darcet, Maxime Oquab, Julien Mairal, and Piotr Bojanowski. 2023.
\newblock Vision transformers need registers.
\newblock \emph{arXiv preprint arXiv:2309.16588}.

\bibitem[{Dehghani et~al.(2018)Dehghani, Gouws, Vinyals, Uszkoreit, and Kaiser}]{dehghani2018universal}
Mostafa Dehghani, Stephan Gouws, Oriol Vinyals, Jakob Uszkoreit, and {\L}ukasz Kaiser. 2018.
\newblock Universal transformers.
\newblock \emph{arXiv preprint arXiv:1807.03819}.

\bibitem[{Dong et~al.(2018)Dong, Quirk, and Lapata}]{dong-etal-2018-confidence}
Li~Dong, Chris Quirk, and Mirella Lapata. 2018.
\newblock \href {https://doi.org/10.18653/v1/P18-1069} {Confidence modeling for neural semantic parsing}.
\newblock In \emph{Proceedings of the 56th Annual Meeting of the Association for Computational Linguistics (Volume 1: Long Papers)}, pages 743--753, Melbourne, Australia. Association for Computational Linguistics.

\bibitem[{Dziri et~al.(2024)Dziri, Lu, Sclar, Li, Jiang, Lin, Welleck, West, Bhagavatula, Le~Bras et~al.}]{dziri2024faith}
Nouha Dziri, Ximing Lu, Melanie Sclar, Xiang~Lorraine Li, Liwei Jiang, Bill~Yuchen Lin, Sean Welleck, Peter West, Chandra Bhagavatula, Ronan Le~Bras, et~al. 2024.
\newblock Faith and fate: Limits of transformers on compositionality.
\newblock \emph{Advances in Neural Information Processing Systems}, 36.

\bibitem[{Faiz et~al.(2023)Faiz, Kaneda, Wang, Osi, Sharma, Chen, and Jiang}]{faiz2023llmcarbon}
Ahmad Faiz, Sotaro Kaneda, Ruhan Wang, Rita~Chukwunyere Osi, Prateek Sharma, Fan Chen, and Lei Jiang. 2023.
\newblock Llmcarbon: Modeling the end-to-end carbon footprint of large language models.
\newblock In \emph{The Twelfth International Conference on Learning Representations}.

\bibitem[{Fernando and Tsokos(2021)}]{fernando2021dynamically}
K~Ruwani~M Fernando and Chris~P Tsokos. 2021.
\newblock Dynamically weighted balanced loss: class imbalanced learning and confidence calibration of deep neural networks.
\newblock \emph{IEEE Transactions on Neural Networks and Learning Systems}, 33(7):2940--2951.

\bibitem[{Freitag and Al-Onaizan(2017)}]{freitag-al-onaizan-2017-beam}
Markus Freitag and Yaser Al-Onaizan. 2017.
\newblock \href {https://doi.org/10.18653/v1/W17-3207} {Beam search strategies for neural machine translation}.
\newblock In \emph{Proceedings of the First Workshop on Neural Machine Translation}, pages 56--60, Vancouver. Association for Computational Linguistics.

\bibitem[{Germann(2003)}]{germann-2003-greedy}
Ulrich Germann. 2003.
\newblock \href {https://aclanthology.org/N03-1010} {Greedy decoding for statistical machine translation in almost linear time}.
\newblock In \emph{Proceedings of the 2003 Human Language Technology Conference of the North {A}merican Chapter of the Association for Computational Linguistics}, pages 72--79.

\bibitem[{Goyal et~al.(2024)Goyal, Ji, Rawat, Menon, Kumar, and Nagarajan}]{goyal2024think}
Sachin Goyal, Ziwei Ji, Ankit~Singh Rawat, Aditya~Krishna Menon, Sanjiv Kumar, and Vaishnavh Nagarajan. 2024.
\newblock \href {https://openreview.net/forum?id=ph04CRkPdC} {Think before you speak: Training language models with pause tokens}.
\newblock In \emph{The Twelfth International Conference on Learning Representations}.

\bibitem[{Herel and Mikolov(2024)}]{herel2024thinking}
David Herel and Tomas Mikolov. 2024.
\newblock Thinking tokens for language modeling.
\newblock \emph{arXiv preprint arXiv:2405.08644}.

\bibitem[{Ho et~al.(2022)Ho, Schmid, and Yun}]{ho2022large}
Namgyu Ho, Laura Schmid, and Se-Young Yun. 2022.
\newblock Large language models are reasoning teachers.
\newblock \emph{arXiv preprint arXiv:2212.10071}.

\bibitem[{Jiang et~al.(2020)Jiang, Wolf, Monz, and de~Rijke}]{jiang2020tldr}
Shaojie Jiang, Thomas Wolf, Christof Monz, and Maarten de~Rijke. 2020.
\newblock Tldr: token loss dynamic reweighting for reducing repetitive utterance generation.
\newblock \emph{arXiv preprint arXiv:2003.11963}.

\bibitem[{Jiang et~al.(2023)Jiang, Xu, Gao, Sun, Liu, Dwivedi-Yu, Yang, Callan, and Neubig}]{jiang-etal-2023-active}
Zhengbao Jiang, Frank Xu, Luyu Gao, Zhiqing Sun, Qian Liu, Jane Dwivedi-Yu, Yiming Yang, Jamie Callan, and Graham Neubig. 2023.
\newblock \href {https://doi.org/10.18653/v1/2023.emnlp-main.495} {Active retrieval augmented generation}.
\newblock In \emph{Proceedings of the 2023 Conference on Empirical Methods in Natural Language Processing}, pages 7969--7992, Singapore. Association for Computational Linguistics.

\bibitem[{Kadavath et~al.(2022)Kadavath, Conerly, Askell, Henighan, Drain, Perez, Schiefer, Hatfield-Dodds, DasSarma, Tran-Johnson et~al.}]{kadavath2022language}
Saurav Kadavath, Tom Conerly, Amanda Askell, Tom Henighan, Dawn Drain, Ethan Perez, Nicholas Schiefer, Zac Hatfield-Dodds, Nova DasSarma, Eli Tran-Johnson, et~al. 2022.
\newblock Language models (mostly) know what they know.
\newblock \emph{arXiv preprint arXiv:2207.05221}.

\bibitem[{Kaplan et~al.(2020)Kaplan, McCandlish, Henighan, Brown, Chess, Child, Gray, Radford, Wu, and Amodei}]{kaplan2020scaling}
Jared Kaplan, Sam McCandlish, Tom Henighan, Tom~B Brown, Benjamin Chess, Rewon Child, Scott Gray, Alec Radford, Jeffrey Wu, and Dario Amodei. 2020.
\newblock Scaling laws for neural language models.
\newblock \emph{arXiv preprint arXiv:2001.08361}.

\bibitem[{Keskar et~al.(2019)Keskar, McCann, Varshney, Xiong, and Socher}]{keskarCTRL2019}
Nitish~Shirish Keskar, Bryan McCann, Lav Varshney, Caiming Xiong, and Richard Socher. 2019.
\newblock {CTRL - A Conditional Transformer Language Model for Controllable Generation}.
\newblock \emph{arXiv preprint arXiv:1909.05858}.

\bibitem[{Leviathan et~al.(2023)Leviathan, Kalman, and Matias}]{leviathan2023fast}
Yaniv Leviathan, Matan Kalman, and Yossi Matias. 2023.
\newblock Fast inference from transformers via speculative decoding.
\newblock In \emph{International Conference on Machine Learning}, pages 19274--19286. PMLR.

\bibitem[{Lin et~al.(2017)Lin, Goyal, Girshick, He, and Doll{\'a}r}]{lin2017focal}
Tsung-Yi Lin, Priya Goyal, Ross Girshick, Kaiming He, and Piotr Doll{\'a}r. 2017.
\newblock Focal loss for dense object detection.
\newblock In \emph{Proceedings of the IEEE international conference on computer vision}, pages 2980--2988.

\bibitem[{Ling et~al.(2017)Ling, Yogatama, Dyer, and Blunsom}]{ling-etal-2017-program}
Wang Ling, Dani Yogatama, Chris Dyer, and Phil Blunsom. 2017.
\newblock \href {https://doi.org/10.18653/v1/P17-1015} {Program induction by rationale generation: Learning to solve and explain algebraic word problems}.
\newblock In \emph{Proceedings of the 55th Annual Meeting of the Association for Computational Linguistics (Volume 1: Long Papers)}, pages 158--167, Vancouver, Canada. Association for Computational Linguistics.

\bibitem[{Mahabadi et~al.(2020)Mahabadi, Belinkov, and Henderson}]{karimi-mahabadi-etal-2020-end}
Rabeeh~Karimi Mahabadi, Yonatan Belinkov, and James Henderson. 2020.
\newblock \href {https://doi.org/10.18653/v1/2020.acl-main.769} {End-to-end bias mitigation by modelling biases in corpora}.
\newblock In \emph{Proceedings of the 58th Annual Meeting of the Association for Computational Linguistics}, pages 8706--8716, Online. Association for Computational Linguistics.

\bibitem[{Microsoft(2023)}]{microsoft_research_2023}
Microsoft. 2023.
\newblock \href {https://www.microsoft.com/en-us/research/blog/phi-2-the-surprising-power-of-small-language-models/} {Phi-2: The surprising power of small language models}.
\newblock [Blog post].

\bibitem[{Mukhoti et~al.(2020)Mukhoti, Kulharia, Sanyal, Golodetz, Torr, and Dokania}]{mukhoti2020calibrating}
Jishnu Mukhoti, Viveka Kulharia, Amartya Sanyal, Stuart Golodetz, Philip Torr, and Puneet Dokania. 2020.
\newblock Calibrating deep neural networks using focal loss.
\newblock \emph{Advances in Neural Information Processing Systems}, 33:15288--15299.

\bibitem[{Nguyen and O{'}Connor(2015)}]{nguyen-oconnor-2015-posterior}
Khanh Nguyen and Brendan O{'}Connor. 2015.
\newblock \href {https://doi.org/10.18653/v1/D15-1182} {Posterior calibration and exploratory analysis for natural language processing models}.
\newblock In \emph{Proceedings of the 2015 Conference on Empirical Methods in Natural Language Processing}, pages 1587--1598, Lisbon, Portugal. Association for Computational Linguistics.

\bibitem[{Radford and Narasimhan(2018)}]{Radford2018ImprovingLU}
Alec Radford and Karthik Narasimhan. 2018.
\newblock \href {https://api.semanticscholar.org/CorpusID:49313245} {Improving language understanding by generative pre-training}.

\bibitem[{Radford et~al.(2018)Radford, Narasimhan, Salimans, Sutskever et~al.}]{radford2018improving}
Alec Radford, Karthik Narasimhan, Tim Salimans, Ilya Sutskever, et~al. 2018.
\newblock Improving language understanding by generative pre-training.

\bibitem[{Rao et~al.(2021)Rao, Zhao, Liu, Lu, Zhou, and Hsieh}]{rao2021dynamicvit}
Yongming Rao, Wenliang Zhao, Benlin Liu, Jiwen Lu, Jie Zhou, and Cho-Jui Hsieh. 2021.
\newblock Dynamicvit: Efficient vision transformers with dynamic token sparsification.
\newblock \emph{Advances in neural information processing systems}, 34:13937--13949.

\bibitem[{Samsi et~al.(2023)Samsi, Zhao, McDonald, Li, Michaleas, Jones, Bergeron, Kepner, Tiwari, and Gadepally}]{samsi2023words}
Siddharth Samsi, Dan Zhao, Joseph McDonald, Baolin Li, Adam Michaleas, Michael Jones, William Bergeron, Jeremy Kepner, Devesh Tiwari, and Vijay Gadepally. 2023.
\newblock From words to watts: Benchmarking the energy costs of large language model inference.
\newblock In \emph{2023 IEEE High Performance Extreme Computing Conference (HPEC)}, pages 1--9. IEEE.

\bibitem[{Shridhar et~al.(2022)Shridhar, Stolfo, and Sachan}]{shridhar2022distilling}
Kumar Shridhar, Alessandro Stolfo, and Mrinmaya Sachan. 2022.
\newblock Distilling reasoning capabilities into smaller language models.
\newblock \emph{arXiv preprint arXiv:2212.00193}.

\bibitem[{Sukhbaatar et~al.(2019)Sukhbaatar, Grave, Lample, Jegou, and Joulin}]{sukhbaatar2019augmenting}
Sainbayar Sukhbaatar, Edouard Grave, Guillaume Lample, Herve Jegou, and Armand Joulin. 2019.
\newblock Augmenting self-attention with persistent memory.
\newblock \emph{arXiv preprint arXiv:1907.01470}.

\bibitem[{Tao et~al.(2023)Tao, Dong, and Xu}]{tao2023dual}
Linwei Tao, Minjing Dong, and Chang Xu. 2023.
\newblock Dual focal loss for calibration.
\newblock In \emph{International Conference on Machine Learning}, pages 33833--33849. PMLR.

\bibitem[{Touvron et~al.(2023)Touvron, Martin, Stone, Albert, Almahairi, Babaei, Bashlykov, Batra, Bhargava, Bhosale et~al.}]{touvron2023llama}
Hugo Touvron, Louis Martin, Kevin Stone, Peter Albert, Amjad Almahairi, Yasmine Babaei, Nikolay Bashlykov, Soumya Batra, Prajjwal Bhargava, Shruti Bhosale, et~al. 2023.
\newblock Llama 2: Open foundation and fine-tuned chat models.
\newblock \emph{arXiv preprint arXiv:2307.09288}.

\bibitem[{Varshney et~al.(2022)Varshney, Luo, and Baral}]{varshney2022can}
Neeraj Varshney, Man Luo, and Chitta Baral. 2022.
\newblock Can open-domain qa reader utilize external knowledge efficiently like humans?
\newblock \emph{arXiv preprint arXiv:2211.12707}.

\bibitem[{Vaswani et~al.(2017)Vaswani, Shazeer, Parmar, Uszkoreit, Jones, Gomez, Kaiser, and Polosukhin}]{vaswani2017attention}
Ashish Vaswani, Noam Shazeer, Niki Parmar, Jakob Uszkoreit, Llion Jones, Aidan~N Gomez, {\L}ukasz Kaiser, and Illia Polosukhin. 2017.
\newblock Attention is all you need.
\newblock \emph{Advances in neural information processing systems}, 30.

\bibitem[{Wang et~al.(2022)Wang, Durmus, Goodman, and Hashimoto}]{wang2022language}
Rose~E Wang, Esin Durmus, Noah Goodman, and Tatsunori Hashimoto. 2022.
\newblock Language modeling via stochastic processes.
\newblock \emph{arXiv preprint arXiv:2203.11370}.

\bibitem[{Wei et~al.(2022)Wei, Wang, Schuurmans, Bosma, Xia, Chi, Le, Zhou et~al.}]{wei2022chain}
Jason Wei, Xuezhi Wang, Dale Schuurmans, Maarten Bosma, Fei Xia, Ed~Chi, Quoc~V Le, Denny Zhou, et~al. 2022.
\newblock Chain-of-thought prompting elicits reasoning in large language models.
\newblock \emph{Advances in neural information processing systems}, 35:24824--24837.

\bibitem[{Wu et~al.(2024)Wu, Jiang, Jiang, He, Luo, Wang, Liu, and Wang}]{wu2024noiseboost}
Kai Wu, Boyuan Jiang, Zhengkai Jiang, Qingdong He, Donghao Luo, Shengzhi Wang, Qingwen Liu, and Chengjie Wang. 2024.
\newblock Noiseboost: Alleviating hallucination with noise perturbation for multimodal large language models.
\newblock \emph{arXiv preprint arXiv:2405.20081}.

\bibitem[{Xue et~al.(2023)Xue, Likhosherstov, Arnab, Houlsby, Dehghani, and You}]{xue2023adaptive}
Fuzhao Xue, Valerii Likhosherstov, Anurag Arnab, Neil Houlsby, Mostafa Dehghani, and Yang You. 2023.
\newblock Adaptive computation with elastic input sequence.
\newblock In \emph{International Conference on Machine Learning}, pages 38971--38988. PMLR.

\end{thebibliography}

\onecolumn

\appendix
\section{Details of Datasets}
\label{apdx:datasets}
We download datasets from Hugging Face repository\footnote{\url{https://huggingface.co/datasets/openai/gsm8k}}\footnote{\url{https://huggingface.co/datasets/deepmind/aqua_rat}}\footnote{\url{https://huggingface.co/datasets/google-research-datasets/mbpp}}. Information regarding the splits and numbers of the data we used is provided in Table \ref{tab:datasets}. For the AQUA-RAT, the original dataset contains over 98k instances, with relatively small validation and test sets of around 200 each. For more efficient experimentation, we sample this data to approximately 74k instances. And due to the limited number of validation and test samples for AQUA-RAT (254 each), we supplement these sets by randomly extracting 2000 samples from the training set. Additionally, to ensure accurate evaluation of the model, we modify the AQUA-RAT options by appending four hashtags (\#\#\#\#) and the correct alphabet at the end of the data, similar to GSM8K, to facilitate learning.

\begin{table}
\centering
\begin{tabular}{cccc}
\toprule
\textbf{Dataset} & \textbf{GSM8K}  & \textbf{AQUA-RAT} & \textbf{MBPP} \\ 
\toprule
\textbf{split} & \multicolumn{3}{c}{\textbf{number of datasets}} \\
\toprule
train & 5,610 & 70,104 &  282\\
validation & - & 2,254 & 90 \\
test & 1,319 & 2,254 & 500 \\
\bottomrule
\end{tabular}
\caption{These are the specific numbers of data involved in the entire experiments. }
\label{tab:datasets}
\end{table}

\section{A Priori Experiments}
\label{apdx:apriori}
In this section, we describe the preliminary experiments conducted prior to our main experiments. Our primary experiments focus on the GSM8K dataset. Initially, we have fine-tuned a model on GSM8K to generate predictions on the test set while also calculating the log probabilities of each token. We have observed that certain parts of the data such as the start and end positions of equations and the points where sentences change consistently exhibit low log probabilities. By identifying these positions, we specify the rule-based approach for inserting \texttt{[PAUSE]} tokens. Subsequently, we insert \texttt{[PAUSE]} tokens at these annotated positions and conduct fine-tuning. The results of these experiments are presented in Table \ref{tab:apriori}.

\paragraph{Glossary of Terms for A Priori Experiments}
\begin{itemize}
\item{Math Front GSM8K (\textbf{MF-G})} : Position \texttt{[PAUSE]} token between equal sign(`=') in math equation and start sign of calculator annotations(`<<') at GSM8K dataset.
\item{Math Behind GSM8K (\textbf{MB-G})} : Position \texttt{[PAUSE]} token behind of end sign of calculator annotations(`>>') at GSM8K dataset.
\item{Sentence Behind (\textbf{SB})} : Position \texttt{[PAUSE]} token right after the sentence in reasoning steps ends.
\end{itemize}

\begin{table}[h]
    \centering
    \begin{tabular}{cl|c}
    \toprule
    \textbf{Base Model} & \textbf{Method} & \textbf{GSM8K} \\ 
    \midrule
    \multirow{5}{*}{Phi-2} &  SFT & 52.08  \\
                        & MF-G & \underline{53.90} \\
                        & MB-G & 52.53  \\
                        & SB & 53.82  \\
                        & DIT(learnable) & 47.15 \\
                        & \textbf{DIT(ours)} & \textbf{56.33}  \\
    \bottomrule
    \end{tabular}
    \caption{Results of comparing the heuristic methods for finding locations to insert the \texttt{[PAUSE]} tokens and DIT. For each method, the best score is indicated in \textbf{bold}, and the second-best score is \underline{underlined}.}
    \label{tab:apriori}
\end{table}

At this point, we consider the \texttt{[PAUSE]} token as a learnable token included in the loss function and proceed with our experiments accordingly. All methods in Table \ref{tab:apriori}, except for DIT (ours), are conducted using this approach. Additionally, we insert ten \texttt{[PAUSE]} tokens at each position, while inserting five \texttt{[PAUSE]} tokens in DIT (learnable) and DIT (ours). However, through additional experiments, we have discovered that excluding the \texttt{[PAUSE]} tokens from the loss calculation is optimal, thereby strongly assigning it the role of a dummy token results in our final experimental setup (DIT) exhibiting probability robustness and improves performance across the overall dataset and models.

\newsavebox{\dit}
\begin{lrbox}{\dit}
\begin{minipage}{\linewidth}
\begin{lstlisting}[language=Python]
def find_Min_Sum(num):(*@\hl{[PAUSE]}@*)
    sum = 0
    i = 2
    while(i * i <= num):(*@\hl{[PAUSE]}@*)
        while(num % i == 0):
            sum += i(*@\hl{[PAUSE]}@*)
            num /= i(*@\hl{[PAUSE]}@*)
        i += 1
    sum += num(*@\hl{[PAUSE]}@*)
    return sum
\end{lstlisting}
\end{minipage}
\end{lrbox}

\newsavebox{\fine}
\begin{lrbox}{\fine}
\begin{minipage}{\linewidth}
\begin{lstlisting}[language=Python]
def find_Min_Sum(num): 
    sum = 0
    i = 2
    while(i * i <= num): 
        while(num % i == 0): 
            sum += i 
            num /= i 
        i += 1
    sum += num 
    return sum
\end{lstlisting}
\end{minipage}
\end{lrbox}

\newsavebox{\ran}
\begin{lrbox}{\ran}
\begin{minipage}{\linewidth}
\begin{lstlisting}[language=Python]
def find_Min_Sum(num)(*@\hl{[PAUSE]}@*):(*@\hl{[PAUSE]}@*)
    sum = 0
    i = 2
    while(i * (*@\hl{[PAUSE]}@*)i (*@\hl{[PAUSE]}@*)<= num): 
        while(num % i == 0): 
            sum += i 
            num /= i 
        i += 1
    sum += num(*@\hl{[PAUSE]}@*) 
    return sum
\end{lstlisting}
\end{minipage}
\end{lrbox}

\newsavebox{\aaw}
\begin{lrbox}{\aaw}
\begin{minipage}{\linewidth}
\begin{lstlisting}[language=Python]
def (*@\hl{[PAUSE]}@*)find_Min_Sum(num):(*@\hl{[PAUSE]}@*)
    sum(*@\hl{[PAUSE]}@*) =(*@\hl{[PAUSE]}@*) 0(*@\hl{[PAUSE]}@*)
    i(*@\hl{[PAUSE]}@*) =(*@\hl{[PAUSE]}@*) 2(*@\hl{[PAUSE]}@*)
    while(i(*@\hl{[PAUSE]}@*) *(*@\hl{[PAUSE]}@*) i(*@\hl{[PAUSE]}@*) <=(*@\hl{[PAUSE]}@*) num):(*@\hl{[PAUSE]}@*) 
        while(num(*@\hl{[PAUSE]}@*) %(*@\hl{[PAUSE]}@*) i(*@\hl{[PAUSE]}@*) ==(*@\hl{[PAUSE]}@*) 0):(*@\hl{[PAUSE]}@*) 
            sum(*@\hl{[PAUSE]}@*) +=(*@\hl{[PAUSE]}@*) i(*@\hl{[PAUSE]}@*) 
            num(*@\hl{[PAUSE]}@*) /=(*@\hl{[PAUSE]}@*) i(*@\hl{[PAUSE]}@*) 
        i(*@\hl{[PAUSE]}@*) +=(*@\hl{[PAUSE]}@*) 1(*@\hl{[PAUSE]}@*)
    sum(*@\hl{[PAUSE]}@*) +=(*@\hl{[PAUSE]}@*) num(*@\hl{[PAUSE]}@*) 
    return(*@\hl{[PAUSE]}@*) sum
\end{lstlisting}
\end{minipage}
\end{lrbox}

\section{Experimental Settings}
\label{apdx:settings}

We conduct experiments on the GSM8K, AQUA-RAT, and MBPP tasks. For each type of \texttt{[PAUSE]} setting, we insert five \texttt{[PAUSE]} tokens into the sequence ($M_{DIT}=5$), with the exception of the original fine-tuning baseline (SFT). The losses associated with predicting \texttt{[PAUSE]} tokens are excluded in all methods. We use learning rates of 1e-5 for GSM8K, and 3e-5 for both the MBPP and AQUA-RAT datasets. Although performance varies with different learning rates, the overall trends remain consistent. Each task is trained for three epochs, using the model with the lowest evaluation loss for inference. Fine-tuning is performed using two RTX 6000 GPUs for GSM8K and MBPP, and two A100 GPUs for AQUA-RAT, with a training and test batch size of 4. 

\section{Training Complexity Notation}
\label{apdx:training_complexity}

Let \(F\) denote the computational cost of a single forward pass, and \(B\) denote the computational cost of a single backward pass (including parameter updates).  

\paragraph{Standard SFT}  
In standard supervised fine-tuning (SFT), each iteration performs one forward pass and one backward pass, yielding an approximate cost of
\[
  \mathrm{Cost}_{\mathrm{SFT}} \approx F + B.
\]

\paragraph{DIT Approach}  
In our DIT approach, we first perform an extra forward pass to identify low log-likelihood positions for inserting \texttt{[PAUSE]} tokens, and then perform a second forward pass together with a backward pass. Concretely, as shown in Figure~\ref{fig:train_img},
\begin{enumerate}
  \item Forward pass(blue line): identify tokens/positions with low log-likelihood. In this step, a single forward pass is performed.
  \item Training after token insertion(green line): train on the augmented input containing \texttt{[PAUSE]} tokens. In this step, a single forward pass followed by a backward pass is performed.
\end{enumerate}
Thus, each DIT iteration incurs a cost of
\[
  \mathrm{Cost}_{\mathrm{DIT}} \approx 2F + B.
\]

The primary cost difference between the standard SFT approach and the proposed DIT method lies in the additional forward pass introduced by DIT. This extra computation serves as a minor cost paid for improved performance. Notably, forward passes are generally less computationally expensive than backward passes and offer greater room for optimization \cite{dao2024flashattention}, suggesting that the performance gain can outweigh the added cost.

\newpage

\section{Examples of Methods}
\label{apdx:example}
We provide examples of the dataset with the \texttt{[PAUSE]} token used during training and inference to illustrate the application of each method. From Table \ref{tab:dit_ex} to Table \ref{tab:aaw_ex}, we extract actual examples from each dataset to visualize how and where the \texttt{[PAUSE]} tokens are inserted.

\subsection{DIT}

\begin{table}[!htbp]
    \centering
    \resizebox{\textwidth}{!}{%
    \begin{tabular}{p{0.15\textwidth} | p{0.80\textwidth}}
    \toprule
    \textbf{Dataset} & \textbf{Example} \\ 
    \midrule
    GSM8K & Instruct: Janet's ducks lay 16 eggs per day. She eats three for breakfast every morning and bakes muffins for her friends every day with four. She sells the remainder at the farmers' market daily for \$2 per fresh duck egg. How much in dollars does she make every day at the farmers' market? \newline
    Output:Jan\hl{\texttt{[PAUSE]}}et sells\hl{\texttt{[PAUSE]}} 16 - 3 - 4 = <<16-3-4=9>>9 duck eggs a day.\hl{\texttt{[PAUSE]}} \newline
    She makes 9 * 2 = \$<<9*2=18>>18 every day at the farmer's market.\hl{\texttt{[PAUSE]}}
    \hl{\texttt{[PAUSE]}}\#\#\#\#\hl{\texttt{[PAUSE]}} 18 \\
    
    \midrule
    AQUA-RAT & Instruct:Question : If $A=x^4 + y^4 = 100$, then the greatest possible value of $x$ is between \newline
    Options : A)0 and 3 B)3 and 6 C)6 and 9 D)9 and 12 E)12 and 15 \newline
    Output:My attempt: \newline
    If\hl{\texttt{[PAUSE]}} $A=x^4+y^4=100$, then the greatest possible value of $x$ would be when $y$ is minimum. \newline
    Let\hl{\texttt{[PAUSE]}} $y^4$ be 0. Now $x^4 = 100$. $x$ should be definitely greater than 3 but less than 4. The only option that fits this range is B \newline
    \hl{\texttt{[PAUSE]}}Hence answer is --\hl{\texttt{[PAUSE]}} \newline
    b) 3 and 6.\hl{\texttt{[PAUSE]}} \#\#\#\# B\\
    
    \midrule
    MBPP & Instruct:Write a python function to find minimum sum of factors of a given number. Output:\newline \usebox{\dit} \\ 
    \bottomrule
    \end{tabular}
    }
    \caption{These are examples of \texttt{[PAUSE]} token insertion in each dataset using the DIT method.}
    \label{tab:dit_ex}
\end{table}

\newpage

\subsection{SFT}

\begin{table}[!htbp]
    \centering
    \resizebox{\textwidth}{!}{%
    \begin{tabular}{p{0.15\textwidth} | p{0.80\textwidth}}
    \toprule
    \textbf{Dataset} & \textbf{Example} \\ 
    \midrule
    GSM8K & Instruct: Janet's ducks lay 16 eggs per day. She eats three for breakfast every morning and bakes muffins for her friends every day with four. She sells the remainder at the farmers' market daily for \$2 per fresh duck egg. How much in dollars does she make every day at the farmers' market? \newline
    Output:Janet sells 16 - 3 - 4 = <<16-3-4=9>>9 duck eggs a day. \newline
    She makes 9 * 2 = \$<<9*2=18>>18 every day at the farmer's market.\newline
    \#\#\#\# 18 \\
    
    \midrule
    AQUA-RAT & Instruct:Question : If $A=x^4 + y^4 = 100$, then the greatest possible value of $x$ is between \newline
    Options : A)0 and 3 B)3 and 6 C)6 and 9 D)9 and 12 E)12 and 15 \newline
    Output:My attempt: \newline
    If $A=x^4+y^4=100$, then the greatest possible value of $x$ would be when $y$ is minimum. \newline
    Let $y^4$ be 0. Now $x^4 = 100$. $x$ should be definitely greater than 3 but less than 4. The only option that fits this range is B \newline
    Hence answer is -- \newline
    b) 3 and 6. \#\#\#\# B\\
    
    \midrule
    MBPP & Instruct:Write a python function to find minimum sum of factors of a given number. Output:\newline \usebox{\fine}  \\
    \bottomrule
    \end{tabular}
    }
    \caption{These are examples of \texttt{[PAUSE]} token insertion in each dataset using the SFT method.}
    \label{tab:fine_ex}
\end{table}

\newpage
\subsection{RAN}

\begin{table}[!htbp]
    \centering
    \resizebox{\textwidth}{!}{%
    \begin{tabular}{p{0.15\textwidth} | p{0.80\textwidth}}
    \toprule
    \textbf{Dataset} & \textbf{Example} \\ 
    \midrule
    GSM8K & Instruct: Janet's ducks lay 16 eggs per day. She eats three for breakfast every morning and bakes muffins for her friends every day with four. She sells the remainder at the farmers' market daily for \$2 per fresh duck egg. How much in dollars does she make every day at the farmers' market? \newline
    Output:Janet sells 16 - 3 - 4 \hl{\texttt{[PAUSE]}}=\hl{\texttt{[PAUSE]}} <<16-3-4=9>\hl{\texttt{[PAUSE]}}>9 duck eggs a \hl{\texttt{[PAUSE]}}day. \newline
    She makes 9 * 2 = \$<<9*2=18\hl{\texttt{[PAUSE]}}>>18 every day at the farmer's market.\newline
    \#\#\#\# 18 \\
    
    \midrule
    AQUA-RAT & Instruct:Question : If $A=x^4 + y^4 = 100$, then the greatest possible value of $x$ is between \newline
    Options : A)0 and 3 B)3 and 6 C)6 and 9 D)9 and 12 E)12 and 15 \newline
    Output:M\hl{\texttt{[PAUSE]}}y attempt: \newline
    If $A=x^4+y^4=100$, then the greatest possible value of \hl{\texttt{[PAUSE]}}$x$ would be when $y$ is minimum. \newline
    Let $y^4$ be 0. Now \hl{\texttt{[PAUSE]}}$x^4 = 100$. $x$ sho\hl{\texttt{[PAUSE]}}uld be definitely greater than 3 but less than 4. The only optio\hl{\texttt{[PAUSE]}}n that fits this range is B \newline
    Hence answer is -- \newline
    b) 3 and 6. \#\#\#\# B\\
    
    \midrule
    MBPP & Instruct:Write a python function to find minimum sum of factors of a given number. Output:\newline \usebox{\ran}  \\
    \bottomrule
    \end{tabular}
    }
    \caption{These are examples of \texttt{[PAUSE]} token insertion in each dataset using the RAN method.}
    \label{tab:ran_ex}
\end{table}

\newpage
\subsection{APPD}

\begin{table}[!htbp]
    \centering
    \resizebox{\textwidth}{!}{%
    \begin{tabular}{p{0.15\textwidth} | p{0.80\textwidth}}
    \toprule
    \textbf{Dataset} & \textbf{Example} \\ 
    \midrule
    GSM8K & Instruct: Janet's ducks lay 16 eggs per day. She eats three for breakfast every morning and bakes muffins for her friends every day with four. She sells the remainder at the farmers' market daily for \$2 per fresh duck egg. How much in dollars does she make every day at the farmers' market? \newline
    Output:\hl{\texttt{[PAUSE]}}\hl{\texttt{[PAUSE]}}\hl{\texttt{[PAUSE]}}\hl{\texttt{[PAUSE]}}\hl{\texttt{[PAUSE]}} Janet sells 16 - 3 - 4 = <<16-3-4=9>>9 duck eggs a day. \newline
    She makes 9 * 2 = \$<<9*2=18>>18 every day at the farmer's market.\newline
    \#\#\#\# 18 \\
    
    \midrule
    AQUA-RAT & Instruct:Question : If $A=x^4 + y^4 = 100$, then the greatest possible value of $x$ is between \newline
    Options : A)0 and 3 B)3 and 6 C)6 and 9 D)9 and 12 E)12 and 15 \newline
    Output:\hl{\texttt{[PAUSE]}}\hl{\texttt{[PAUSE]}}\hl{\texttt{[PAUSE]}}\hl{\texttt{[PAUSE]}}\hl{\texttt{[PAUSE]}} My attempt: \newline
    If $A=x^4+y^4=100$, then the greatest possible value of $x$ would be when $y$ is minimum. \newline
    Let $y^4$ be 0. Now $x^4 = 100$. $x$ should be definitely greater than 3 but less than 4. The only option that fits this range is B \newline
    Hence answer is -- \newline
    b) 3 and 6. \#\#\#\# B\\
    
    \midrule
    MBPP & Instruct:Write a python function to find minimum sum of factors of a given number. Output:\hl{\texttt{[PAUSE]}}\hl{\texttt{[PAUSE]}}\hl{\texttt{[PAUSE]}}\hl{\texttt{[PAUSE]}}\hl{\texttt{[PAUSE]}}\newline \usebox{\fine}  \\
    \bottomrule
    \end{tabular}
    }
    \caption{These are examples of \texttt{[PAUSE]} token insertion in each dataset using the APPD method.}
    \label{tab:appd_ex}
\end{table}

\newpage
\subsection{AAW}

\begin{table}[!htbp]
    \centering
    \resizebox{\textwidth}{!}{%
    \begin{tabular}{p{0.15\textwidth} | p{0.80\textwidth}}
    \toprule
    \textbf{Dataset} & \textbf{Example} \\ 
    \midrule
    GSM8K & Instruct: Janet's ducks lay 16 eggs per day. She eats three for breakfast every morning and bakes muffins for her friends every day with four. She sells the remainder at the farmers' market daily for \$2 per fresh duck egg. How much in dollars does she make every day at the farmers' market? \newline
    Output:Janet\hl{\texttt{[PAUSE]}} sells\hl{\texttt{[PAUSE]}} 16\hl{\texttt{[PAUSE]}} -\hl{\texttt{[PAUSE]}} 3\hl{\texttt{[PAUSE]}} -\hl{\texttt{[PAUSE]}} 4\hl{\texttt{[PAUSE]}} =\hl{\texttt{[PAUSE]}} <<16-3-4=9>>9\hl{\texttt{[PAUSE]}} duck\hl{\texttt{[PAUSE]}} eggs\hl{\texttt{[PAUSE]}} a\hl{\texttt{[PAUSE]}} day.\hl{\texttt{[PAUSE]}} \newline
    She\hl{\texttt{[PAUSE]}} makes\hl{\texttt{[PAUSE]}} 9\hl{\texttt{[PAUSE]}} *\hl{\texttt{[PAUSE]}} 2\hl{\texttt{[PAUSE]}} =\hl{\texttt{[PAUSE]}} \$<<9*2=18>>18\hl{\texttt{[PAUSE]}} every\hl{\texttt{[PAUSE]}} day\hl{\texttt{[PAUSE]}} at\hl{\texttt{[PAUSE]}} the\hl{\texttt{[PAUSE]}} farmer's\hl{\texttt{[PAUSE]}} market.\hl{\texttt{[PAUSE]}}\newline
    \#\#\#\#\hl{\texttt{[PAUSE]}} 18 \\
    
    \midrule
    AQUA-RAT & Instruct:Question : If $A=x^4 + y^4 = 100$, then the greatest possible value of $x$ is between \newline
    Options : A)0 and 3 B)3 and 6 C)6 and 9 D)9 and 12 E)12 and 15 \newline
    Output:My\hl{\texttt{[PAUSE]}} attempt:\hl{\texttt{[PAUSE]}} \newline
    If\hl{\texttt{[PAUSE]}} $A=x^4+y^4=100$,\hl{\texttt{[PAUSE]}} then\hl{\texttt{[PAUSE]}} the\hl{\texttt{[PAUSE]}} greatest\hl{\texttt{[PAUSE]}} possible\hl{\texttt{[PAUSE]}} value\hl{\texttt{[PAUSE]}} of\hl{\texttt{[PAUSE]}} $x$\hl{\texttt{[PAUSE]}} would\hl{\texttt{[PAUSE]}} be\hl{\texttt{[PAUSE]}} when\hl{\texttt{[PAUSE]}} $y$\hl{\texttt{[PAUSE]}} is\hl{\texttt{[PAUSE]}} minimum.\hl{\texttt{[PAUSE]}} \newline
    Let\hl{\texttt{[PAUSE]}} $y^4$\hl{\texttt{[PAUSE]}} be\hl{\texttt{[PAUSE]}} 0.\hl{\texttt{[PAUSE]}} Now\hl{\texttt{[PAUSE]}} $x^4\hl{\texttt{[PAUSE]}} =\hl{\texttt{[PAUSE]}} 100$.\hl{\texttt{[PAUSE]}} $x$\hl{\texttt{[PAUSE]}} should\hl{\texttt{[PAUSE]}} be\hl{\texttt{[PAUSE]}} definitely\hl{\texttt{[PAUSE]}} greater\hl{\texttt{[PAUSE]}} than\hl{\texttt{[PAUSE]}} 3\hl{\texttt{[PAUSE]}} but\hl{\texttt{[PAUSE]}} less\hl{\texttt{[PAUSE]}} than\hl{\texttt{[PAUSE]}} 4.\hl{\texttt{[PAUSE]}} The\hl{\texttt{[PAUSE]}} only\hl{\texttt{[PAUSE]}} option\hl{\texttt{[PAUSE]}} that\hl{\texttt{[PAUSE]}} fits\hl{\texttt{[PAUSE]}} this\hl{\texttt{[PAUSE]}} range\hl{\texttt{[PAUSE]}} is\hl{\texttt{[PAUSE]}} B\hl{\texttt{[PAUSE]}} \newline
    Hence\hl{\texttt{[PAUSE]}} answer\hl{\texttt{[PAUSE]}} is\hl{\texttt{[PAUSE]}} --\hl{\texttt{[PAUSE]}} \newline
    b)\hl{\texttt{[PAUSE]}} 3\hl{\texttt{[PAUSE]}} and\hl{\texttt{[PAUSE]}} 6.\hl{\texttt{[PAUSE]}} \#\#\#\#\hl{\texttt{[PAUSE]}} B\\
    
    \midrule
    MBPP & Instruct:Write a python function to find minimum sum of factors of a given number. Output:\newline \usebox{\aaw}  \\
    \bottomrule
    \end{tabular}
    }
    \caption{These are examples of \texttt{[PAUSE]} token insertion in each dataset using the AAW method.}
    \label{tab:aaw_ex}
\end{table}

\end{document}